\pdfoutput=1

\documentclass[11pt]{article}  

\usepackage{caption}        
\usepackage{threeparttable} 
\usepackage{array}          
\usepackage{booktabs}       
\usepackage{afterpage}      

\usepackage[final]{acl}

\usepackage{times}
\usepackage{latexsym}

\usepackage[T1]{fontenc}

\usepackage[utf8]{inputenc}

\usepackage{microtype}

\usepackage{inconsolata}

\usepackage{graphicx}

\title{Improving Narrative Classification and Explanation via Fine Tuned Language Models}

\author{
  Rishit Tyagi* \\
  {\normalsize \texttt{tyagirishit21@gmail.com}} \\
  \And
  Rahul Bouri* \\
  {\normalsize \texttt{rahulbouri16@gmail.com}} \\
  \And
  Mohit Gupta* \\
  {\normalsize \texttt{mohit.gupta2002@gmail.com}} \\
}

\begin{document}
\maketitle
\renewcommand{\thefootnote}{\fnsymbol{footnote}}
\footnotetext[1]{These authors contributed equally to this work.}

\begin{abstract}
Understanding covert narratives and implicit messaging is essential for analyzing bias and sentiment. Traditional NLP methods struggle with detecting subtle phrasing and hidden agendas. This study tackles two key challenges: (1) multi-label classification of narratives and sub-narratives in news articles, and (2) generating concise, evidence-based explanations for dominant narratives. We fine-tune a BERT model with a recall-oriented approach for comprehensive narrative detection, refining predictions using a GPT-4o pipeline for consistency. For narrative explanation, we propose a ReACT (Reasoning + Acting) framework with semantic retrieval-based few-shot prompting, ensuring grounded and relevant justifications. To enhance factual accuracy and reduce hallucinations, we incorporate a structured taxonomy table as an auxiliary knowledge base. Our results show that integrating auxiliary knowledge in prompts improves classification accuracy and justification reliability, with applications in media analysis, education, and intelligence gathering.

\end{abstract}

\section{Introduction}

The rise of digital media has dramatically reshaped the way information is produced and consumed, enabling direct communication between content creators and audiences. Although this has democratized information access, it has also made it easier for manipulative narratives and disinformation to spread, especially during crises and politically sensitive events. News articles often employ implicit messaging, strategic framing, and loaded language to subtly shape public perception \cite{Mokhberian_2020}. These covert techniques are not always explicitly deceptive, but instead rely on suggestive phrasing, selective omissions, and emotionally charged language, making them difficult to detect through traditional Natural Language Processing (NLP) methods.

This phenomenon is especially common in geopolitical conflicts and environmental discourse, where language is often used to shape ideological perspectives, downplay motivations, or influence opinions. For example, narratives surrounding climate change policies or the Ukraine-Russia conflict frequently employ carefully constructed rhetoric to promote certain viewpoints without making direct claims. Identifying these hidden patterns is essential to analyze the influence of the media and counter disinformation. Beyond news media, the ability to detect implicit meaning is valuable in various domains such as education, legal analysis, cross-cultural studies and security.

This study builds upon the foundation laid by prior research in implicit narrative detection and develops a system designed to address the objectives and evaluation framework introduced in \cite{semeval2025task10}. Specifically, we focus on two key tasks in the analysis of implicit narratives in news articles. First, we fine-tune bert-base-uncased \cite{devlin2019bert} for the multi-label classification task to identify and categorize dominant narratives present in a given text. This is then passed through a prompt engineered Large Language Model (LLM) to identify the final classification from the shortened classification list returned by BERT. Second, we introduce a methodology for generating structured justifications that explain why a particular narrative has been assigned to a text. This explanation process relies on retrieving semantically relevant evidence from the article itself and structuring the justification using a ReACT (Reasoning + Acting) framework \cite{yao2023react}. To enhance the factual reliability of these justifications, we incorporate a taxonomy-based knowledge lookup, which provides formal definitions and examples of narratives and sub-narratives.

By refining methods for extracting implied meaning, this research contributes to media analysis, automated content understanding, and intelligence gathering. Ultimately, advancing NLP-driven narrative detection will provide deeper insight into how narratives influence perception in diverse languages, cultures, and discourse contexts.

\section{Related Work}

The task of extracting dominant and sub-narratives from text and also generating free-text explanations that justify a dominant narrative within a text article falls under the broader domain of computational narrative extraction and discourse analysis. These are fundamental tasks in NLP and we have explored research on multilabel text classification, Explainable AI (XAI) for text generation, Retrieval-Augmented Generation (RAG), and ReACT prompting techniques.

\subsection{Narrative Extraction}

Narrative extraction tasks have their roots in the extraction of 'topic' and 'event'. \cite{feng2018language} has worked on language-independent neural networks to capture sequence and semantic information for event detection. This approach though multilingual is not effective in extraction of narratives where content is implicit with subtle language dependent paraphrasing causing complex dependencies amongst narratives and sub-narratives.
\newline With advancements in encoder-decoder architectures and the semantic capabilities of large language models (LLMs), significant progress has been made in natural language understanding. However, key challenges remain: (1) designing annotation schemes that are both comprehensive enough to capture narrative features while remaining concise to prevent input dilution and hallucinations \cite{huang2025survey}; (2) ensuring robustness across diverse writing styles (formal/informal) and multilingual inputs \cite{Qin2024MultilingualLL}; and (3) improving explainability in intermediate steps to enhance the interpretability of results \cite{zhao2024explainability}.
Another critical issue in fine-tuning LLMs is data scarcity and class imbalance, which can negatively impact model performance. To address this, ensemble methods have been explored as a way to leverage complementary strengths across models. \cite{randl-etal-2024-cicle} employs such an ensemble-based classification approach, which proves effective for extracting labels when explicit class mentions are present in the text. However, this method has limitations, particularly in handling implicit features and lacks intermediate explanatory steps, which are crucial for improving transparency and interpretability.

\subsection{Narrative Explanation}

Research in narrative explanation is rooted in Explainable AI (XAI) frameworks designed to ensure factual consistency. While limited work exists on multi-label narrative justification, traditional approaches often employ Named Entity Recognition (NER) to model sentence structures \cite{santana2023survey}, integrating these with text generation models to produce coherent outputs. Although computationally efficient, these methods struggle with hierarchical labels, where dominant narratives encompass multiple sub-narratives, leading to subtle modifications in the overall explanation.
Recent advancements in large language models (LLMs), particularly with reasoning-enhancing prompting techniques such as "ReACT" \cite{yao2023react}, "Chain-of-Thought" \cite{wei2022chain}, and "Tree-of-Thought" \cite{yao2023tree}, have demonstrated promising results in structured reasoning. However, these approaches often fail to capture the full complexity of hierarchical relationships, especially when critical information is embedded in short sentences or within the dataset’s taxonomy.

\section{Task Description}

Understanding implicit narratives in news articles is essential for detecting bias, framing, and potential manipulation. This task focuses on two key challenges: multi-label classification of narratives and sub-narratives (Subtask 2) and generating concise, evidence-based explanations for dominant narratives (Subtask 3).
\newline In Subtask 2 (Narrative Classification), given a news article and a predefined two-level taxonomy of narratives and sub-narratives, the goal is to accurately assign all relevant sub-narrative labels to the article. This is a multi-class, multi-label classification problem where both the primary narrative and its sub-narratives must be correctly identified.
\newline In Subtask 3 (Narrative Extraction), given a news article and a dominant narrative, the goal is to generate a brief, text-based explanation (maximum 80 words) supporting the dominant narrative. The generated justification must be grounded in the article by referencing textual evidence that aligns with the claims of the dominant narrative.
Both subtasks are crucial for enhancing media analysis, fact-checking, and disinformation detection by providing structured narrative classification and transparent, text-grounded justifications.

\section{Methodology}

This section describes our approach to solving the two subtasks: (1) Narrative Classification, where we assigned narratives and sub-narratives to news articles in a multi-label classification setup, and (2) Narrative Explanation, where we generated grounded justifications for dominant narratives using a retrieval-augmented LLM-based approach.

\subsection{Narrative Classification (Subtask 2)}

\subsubsection{Data Preparation}

The dataset consisted of news and web articles in five languages (Bulgarian, English, Hindi, Portuguese, and Russian), focusing on the Ukraine-Russia war and climate change. Each article was labeled with a dominant narrative and one or more sub-narratives. We structured the dataset for training by one-hot encoding the dominant narrative labels, enabling a multi-label classification setup. The data was split into an 80-20 train-validation split for training.

\subsubsection{Fine-Tuned BERT Model}

For classification, we fine-tuned a BERT-base-uncased model with focal loss \cite{cao2021adaptable} to address label imbalance implicitly. The model was fine-tuned with a primary focus on maximizing recall to ensure the inclusion of all relevant labels \cite{sun2019fine}. In multi-label classification tasks, precision and recall present a trade-off \cite{zhang2019theoretically}: prioritizing recall increases the likelihood of retrieving all relevant labels, albeit at the cost of increased false positives (Type I errors). Given the hierarchical nature of our approach, where we have a second highly specific classification step, missing a correct label is more detrimental than including an incorrect one (Type II errors). Hence we ensure that relevant labels are not missed by prioritising high recall.
\newline The model was trained for eight epochs with a batch size of 8, a learning rate of 2e-5, and the AdamW optimizer ($\epsilon = 1e-8$, weight decay = 0.05). 
A lower weight decay was used to prevent excessive regularization, which could suppress recall. 
Additionally, a linear learning rate scheduler with a 10\% warm-up was applied to improve convergence stability. 
To further minimise false negatives, we applied adaptive threshold tuning, ensuring that relevant labels were 
retained without excessively increasing false positives. Increasing the decision threshold reduces the risk of Type II errors but raises the likelihood of Type I errors. Given our priority on recall, we adjusted the threshold adaptively to minimize false negatives while maintaining an acceptable false positive rate.
\newline Finally, we trained two separate models: one for Climate Change narratives and another for Ukraine-Russia War 
narratives, ensuring task-specific adaptation.

\subsubsection{GPT-4o Post-Processing}
To refine the BERT predictions, we implemented a two-stage \texttt{GPT-4o} pipeline leveraging taxonomy-based reasoning. We employed Tree-of Thought prompting techniques \cite{yao2023tree} to encourage the Large Language Model to evaluate intermediate steps and solve the problem with a structured reasoning process. The process involved:

\begin{enumerate}
    \item \textbf{Narrative Label Refinement:} The article and initial BERT-predicted labels were passed to \texttt{GPT-4o} along with a taxonomy defining the meaning of each narrative. The model was instructed to filter incorrect labels while ensuring true positives were retained.
    \item \textbf{Sub-Narrative Classification:} Given the refined narrative labels, \texttt{GPT-4o} was prompted again with a taxonomy for sub-narratives corresponding to each narrative, generating the final set of sub-narrative labels.
\end{enumerate}
This approach helped enforce hierarchical label consistency and align predictions with predefined taxonomies.
\newline
\subsection{Narrative Explanation (Subtask 3)}
\subsubsection{Semantic Sentence Retrieval}
For generating evidence-based justifications, we combined semantic sentence retrieval \cite{jingling2014sentence} with GPT-4o based ReACT prompting to ensure explanations were grounded in the article text. Our retrieval approach involved:
\begin{enumerate}
    \item Sentence Segmentation: Articles were split into sentences using period-based segmentation.
    \item Semantic Indexing: Each sentence is embedded using OpenAI’s text-embedding-ada-002 model \cite{rodriguez2022word} and stored in a vector database. Cosine similarity is used as the distance metric for retrieval. After retrieval, the article is deleted from the database to optimize memory usage.
    \item Dual-Pass Cosine Similarity Retrieval: Top 5 sentences were retrieved based on cosine similarity with the dominant narrative. A second retrieval was then performed for sub-narratives, adding any sentence that exceeded the similarity threshold set by the 5th-ranked sentence from the first retrieval.
\end{enumerate}
This dynamic thresholding ensured that only semantically relevant sentences were used while preventing arbitrary cutoff points.

\subsubsection{ReACT-Based Prompting}

To generate structured and interpretable justifications, we implemented a ReACT (Reasoning + Acting) framework that follows a chain-of-thought reasoning process. This approach ensures that explanations are logically structured and grounded in the retrieved text. The process involves three key steps: (1) identifying central claims, (2) justifying the dominant narrative, and (3) justifying the sub-narrative.
First, the model identifies central claims by analyzing the retrieved sentences and detecting references to key themes and implicit messaging. For example, if a text discusses globalists and environmentalists orchestrating events in secret, the model searches for evidence of powerful groups exerting hidden influence, such as mentions of "globalists," "communists," and "environmentalists" manipulating public opinion.
\newline Next, the model justifies the dominant narrative by identifying claims that reinforce the overarching theme. If the dominant narrative suggests that climate policies are part of a coordinated, deceptive effort by powerful entities, the model locates supporting statements, such as assertions that globalists "deliberately start fires" or "use climate change as an excuse for depopulation." Based on this evidence, the model concludes that the dominant narrative aligns with "Hidden plots by secret schemes of powerful groups."
\newline Finally, the model applies the same process to justify the sub-narrative, focusing on more specific underlying themes. If the sub-narrative suggests that climate policies have an ulterior motive beyond environmental concerns, the model extracts relevant claims, such as statements equating sustainability efforts with abortion and depopulation agendas. This leads to the conclusion that the text supports the sub-narrative of "The climate agenda has hidden motives."
\newline To optimize this reasoning process, we experimented with few-shot prompting but found that ReACT prompting yielded more structured and interpretable justifications. By breaking down the process into Thought, Action, Observation, and Conclusion, the model systematically evaluates retrieved evidence, minimizing inconsistencies and improving transparency.

\renewcommand{\tablename}{Table}
\renewcommand{\thetable}{\arabic{table}}
\begin{table}[t]
    \centering
    \captionsetup{font=normalsize}
    \small
    \begin{tabular}{p{3.2cm} p{3.2cm}} 
        \toprule
        \textbf{Column} & \textbf{Description} \\
        \midrule
        Main Narrative & Unique identifier for the dominant narrative. \\
        Main Narrative Definition & Ground-truth definition of the dominant narrative. \\
        Main Narrative Example & Example cases supporting the dominant narrative. \\
        Metadata (Main Narrative) & Additional distinguishing attributes. \\
        Sub-Narrative & Unique identifier for the sub-narrative. \\
        Sub-Narrative Definition & Ground-truth definition of the sub-narrative. \\
        Sub-Narrative Example & Example cases supporting the sub-narrative. \\
        Metadata (Sub-Narrative) & Additional distinguishing attributes. \\
        \bottomrule
    \end{tabular}
    \caption{Narrative Taxonomy Specifications}
    \label{tab:narrative-specs}
\end{table}

\subsubsection{Taxonomy Table Integration}
While prompting and retrieval alone improve justification generation, we introduce a structured taxonomy table as an auxiliary knowledge base to further enhance interpretability and factual alignment. We tested two approaches for integrating this information:
Explicitly inserting the taxonomy table as instructions \cite{sarmah2024hybridrag} in the prompt.
Embedding it within the "Action" section of the ReACT prompt.
Our experiments found that the second approach gave better results, as defining the taxonomy as a part of the Action section led to more reliable and factually consistent justifications.

\renewcommand{\tablename}{Table}
\renewcommand{\thetable}{2}
\setcounter{table}{0}  

\begin{table*}[t]
    \centering
    \captionsetup{font=normalsize}

    \begin{minipage}{\linewidth}
        \centering
        \resizebox{\linewidth}{!}{ 
        \begin{tabular}{cccccc}
            \hline
            Task & GPT 4o-mini & GPT 4o & BERT + GPT 4o-mini & BERT + GPT 4o \\
            \hline
            Narrative CC & 0.227 & 0.227 & 0.6 &  0.6 \\
            Narrative URW & 0.301 & 0.301 & 0.342 & 0.326 \\
            Narrative Overall & 0.251 & 0.251 & 0.458 & \textbf{0.467} \\
            Sub Narrative CC & 0.156 & 0.158 & 0.239 & 0.244 \\
            Sub Narrative URW & 0.187 & 0.187 & 0.188 & 0.2 \\
            Sub Narrative Overall & 0.164 & 0.166 & 0.208 & \textbf{0.217} \\
            \hline
        \end{tabular}
        }
        \caption{Classification F1 Scores}
        \label{tab:first}
    \end{minipage}
\end{table*}

\renewcommand{\tablename}{Table}
\renewcommand{\thetable}{3}
\setcounter{table}{0}  

\begin{table*}[t]
    \centering
    \captionsetup{font=normalsize}

    \begin{minipage}{\linewidth}
        \centering
        \resizebox{\linewidth}{!}{ 
        \begin{tabular}{cccccc}
            \hline
            & BG & EN & HI & PT & RU \\
            \hline
            Simple ReACT Prompt & 0.6018 & 0.618 & 0.6308 & 0.6529 & 0.6252 \\
            ReACT with Auxiliary Knowledge Base & 0.6114 & 0.6288 & 0.6605 & 0.6904 & 0.6374 \\
            ReACT with Auxiliary Knowledge Base and Semantic Search & \textbf{0.6720} & \textbf{0.6910} & \textbf{0.7271} & \textbf{0.7192} & \textbf{0.6644} \\
            \hline
        \end{tabular}
        }
        \caption{BERT Score F1 Results}
        \label{tab:f1-score}
    \end{minipage}
\end{table*}

\section{Results}
We evaluated our two tasks—multilabel classification and the generation of evidence-based explanations for narratives—using F1 scores. The approach with the highest F1 score was chosen as the objective, as we aimed to balance precision and recall while minimizing False Positives and False Negatives.
\newline For the text generation task, we utilized BERTScore \cite{zhang2019bertscore} to compare results, as it measures semantic similarity between strings using contextual embeddings. Unlike n-gram-based metrics (e.g., BLEU, ROUGE) \cite{culy2003limits}, which struggle with paraphrased or implicit reasoning, BERTScore effectively captures meaning equivalence by leveraging deep contextual representations.
\newline Table 2 presents results for Narrative Classification (Subtask 2) across experiments, while Table 3 showcases results for Narrative Justification (Subtask 3).

\subsection{Narrative Classification}
Our framework for predicting dominant and sub-narratives achieved an overall F1 score of 0.467 and 0.217, respectively, when using \texttt{GPT-4o}. The results with \texttt{GPT-4o-mini} were comparable, yielding 0.458 and 0.208 for dominant and sub-narratives, respectively. These findings were compiled after the task was complete, and highlight how refining our approach with a weaker classifier before the final classification step provides better results while keeping the context for the LLM as concise as possible.

\subsection{Narrative Justification}
The text generation resulting from our novel approach—integrating Semantic Similarity Search to retrieve sentences from the article text and pairing them with an auxiliary knowledge base in a ReACT prompt—consistently outperformed both the simple prompt and the prompt paired solely with the knowledge base across all five languages (Bulgarian, English, Hindi, Portuguese, and Russian). Moreover, our text justification framework demonstrated superior performance across all three evaluation metrics—BERT F1, Precision, and Recall—consistently surpassing alternative approaches. These results emphasize the effectiveness of leveraging semantic similarity and external knowledge augmentation to enhance justification quality across multilingual settings. The results prove that the efficiency of the designed framework allows us to utilize smaller LLMs in future work, enhancing scalability.

\section{Conclusion}
This study advances NLP-driven narrative analysis by introducing a framework for classifying and justifying implicit narratives in news articles. Our multilabel classification approach, fine-tuning \texttt{bert-base-uncased} with a prompt-engineered LLM, effectively identified dominant and sub-narratives. The framework maintained strong performance even with \texttt{GPT-4o-mini}, demonstrating the scalability and adaptability of the system without significant performance compromises. This lightweight configuration reduces computational overhead and enables deployment in resource-constrained environments, making the framework practical for real-world, large-scale applications. Future implementations can further optimize resource usage by incorporating retrieval caching mechanisms and distributed modular processing across subtasks.
\newline Furthermore, our narrative justification approach, which combines Semantic Similarity Search with a ReACT-based reasoning structure and auxiliary knowledge retrieval, significantly improved text generation quality across multiple languages. The model consistently outperformed the baseline methods in BERT F1, underscoring the effectiveness of integrating contextual retrieval mechanisms with generative reasoning to generate coherent and factually aligned justifications.
\newline These findings improve media analysis, automated content understanding, and intelligence gathering by improving the detection of implicit ideological framing. As NLP advances, our approach lays the groundwork for more transparent, explainable AI-driven media analysis, supporting efforts to combat misinformation and strengthen media literacy across diverse linguistic and cultural contexts. Looking ahead, future work will investigate the use of dynamic crowd-sourced knowledge bases and adversarial testing to identify and minimize potential biases introduced via semantic retrieval or taxonomy-driven prompting. This will further ensure the fairness, robustness and generalizability of the system across sociopolitical domains and multilingual settings.

\section*{Acknowledgments}

We would like to thank the organisers of \emph{SemEval 2025 Task 10: Multilingual Characterization and Extraction of Narratives from Online News} for encouraging research on such crucial topics.

\bibliography{custom}

\begin{thebibliography}{20}
\providecommand{\natexlab}[1]{#1}

\bibitem[{Cao et~al.(2021)Cao, Liu, and Shen}]{cao2021adaptable}
Lu~Cao, Xinyue Liu, and Hong Shen. 2021.
\newblock Adaptable focal loss for imbalanced text classification.
\newblock In \emph{International Conference on Parallel and Distributed Computing: Applications and Technologies}, pages 466--475. Springer.

\bibitem[{Culy and Riehemann(2003)}]{culy2003limits}
Chris Culy and Susanne~Z Riehemann. 2003.
\newblock The limits of n-gram translation evaluation metrics.
\newblock In \emph{Proceedings of Machine Translation Summit IX: Papers}.

\bibitem[{Devlin et~al.(2019)Devlin, Chang, Lee, and Toutanova}]{devlin2019bert}
Jacob Devlin, Ming-Wei Chang, Kenton Lee, and Kristina Toutanova. 2019.
\newblock Bert: Pre-training of deep bidirectional transformers for language understanding.
\newblock In \emph{Proceedings of the 2019 conference of the North American chapter of the association for computational linguistics: human language technologies, volume 1 (long and short papers)}, pages 4171--4186.

\bibitem[{Feng et~al.(2018)Feng, Qin, and Liu}]{feng2018language}
Xiaocheng Feng, Bing Qin, and Ting Liu. 2018.
\newblock A language-independent neural network for event detection.
\newblock \emph{Science China Information Sciences}, 61:1--12.

\bibitem[{Huang et~al.(2025)Huang, Yu, Ma, Zhong, Feng, Wang, Chen, Peng, Feng, Qin et~al.}]{huang2025survey}
Lei Huang, Weijiang Yu, Weitao Ma, Weihong Zhong, Zhangyin Feng, Haotian Wang, Qianglong Chen, Weihua Peng, Xiaocheng Feng, Bing Qin, et~al. 2025.
\newblock A survey on hallucination in large language models: Principles, taxonomy, challenges, and open questions.
\newblock \emph{ACM Transactions on Information Systems}, 43(2):1--55.

\bibitem[{Jingling et~al.(2014)Jingling, Huiyun, and Baojiang}]{jingling2014sentence}
Zhao Jingling, Zhang Huiyun, and Cui Baojiang. 2014.
\newblock Sentence similarity based on semantic vector model.
\newblock In \emph{2014 Ninth International Conference on P2P, Parallel, Grid, Cloud and Internet Computing}, pages 499--503. IEEE.

\bibitem[{Mokhberian et~al.(2020)Mokhberian, Abeliuk, Cummings, and Lerman}]{Mokhberian_2020}
Negar Mokhberian, Andrés Abeliuk, Patrick Cummings, and Kristina Lerman. 2020.
\newblock \href {https://doi.org/10.1007/978-3-030-60975-7_16} {\emph{Moral Framing and Ideological Bias of News}}, page 206–219.
\newblock Springer International Publishing.

\bibitem[{Piskorski et~al.(2025)Piskorski, Mahmoud, Nikolaidis, Campos, Jorge, Dimitrov, Silvano, Yangarber, Sharma, Chakraborty, Guimarães, Sartori, Stefanovitch, Xie, Nakov, and Da~San~Martino}]{semeval2025task10}
Jakub Piskorski, Tarek Mahmoud, Nikolaos Nikolaidis, Ricardo Campos, Alípio Jorge, Dimitar Dimitrov, Purificação Silvano, Roman Yangarber, Shivam Sharma, Tanmoy Chakraborty, Nuno Guimarães, Elisa Sartori, Nicolas Stefanovitch, Zhuohan Xie, Preslav Nakov, and Giovanni Da~San~Martino. 2025.
\newblock {SemEval}-2025 task 10: Multilingual characterization and extraction of narratives from online news.
\newblock In \emph{Proceedings of the 19th International Workshop on Semantic Evaluation}, SemEval 2025, Vienna, Austria.

\bibitem[{Qin et~al.(2024)Qin, Chen, Zhou, Chen, Li, Liao, Li, Che, and Yu}]{Qin2024MultilingualLL}
Libo Qin, Qiguang Chen, Yuhang Zhou, Zhi Chen, Yinghui Li, Lizi Liao, Min Li, Wanxiang Che, and Philip~S. Yu. 2024.
\newblock \href {https://api.semanticscholar.org/CorpusID:269005862} {Multilingual large language model: A survey of resources, taxonomy and frontiers}.
\newblock \emph{ArXiv}, abs/2404.04925.

\bibitem[{Randl et~al.(2024)Randl, Pavlopoulos, Henriksson, and Lindgren}]{randl-etal-2024-cicle}
Korbinian Randl, John Pavlopoulos, Aron Henriksson, and Tony Lindgren. 2024.
\newblock \href {https://doi.org/10.18653/v1/2024.findings-acl.459} {{CICL}e: Conformal in-context learning for largescale multi-class food risk classification}.
\newblock In \emph{Findings of the Association for Computational Linguistics: ACL 2024}, pages 7695--7715, Bangkok, Thailand. Association for Computational Linguistics.

\bibitem[{Rodriguez and Spirling(2022)}]{rodriguez2022word}
Pedro~L Rodriguez and Arthur Spirling. 2022.
\newblock Word embeddings: What works, what doesn’t, and how to tell the difference for applied research.
\newblock \emph{The Journal of Politics}, 84(1):101--115.

\bibitem[{Santana et~al.(2023)Santana, Campos, Amorim, Jorge, Silvano, and Nunes}]{santana2023survey}
Brenda Santana, Ricardo Campos, Evelin Amorim, Al{\'\i}pio Jorge, Purifica{\c{c}}{\~a}o Silvano, and S{\'e}rgio Nunes. 2023.
\newblock A survey on narrative extraction from textual data.
\newblock \emph{Artificial Intelligence Review}, 56(8):8393--8435.

\bibitem[{Sarmah et~al.(2024)Sarmah, Mehta, Hall, Rao, Patel, and Pasquali}]{sarmah2024hybridrag}
Bhaskarjit Sarmah, Dhagash Mehta, Benika Hall, Rohan Rao, Sunil Patel, and Stefano Pasquali. 2024.
\newblock Hybridrag: Integrating knowledge graphs and vector retrieval augmented generation for efficient information extraction.
\newblock In \emph{Proceedings of the 5th ACM International Conference on AI in Finance}, pages 608--616.

\bibitem[{Sun et~al.(2019)Sun, Qiu, Xu, and Huang}]{sun2019fine}
Chi Sun, Xipeng Qiu, Yige Xu, and Xuanjing Huang. 2019.
\newblock How to fine-tune bert for text classification?
\newblock In \emph{China national conference on Chinese computational linguistics}, pages 194--206. Springer.

\bibitem[{Wei et~al.(2022)Wei, Wang, Schuurmans, Bosma, Xia, Chi, Le, Zhou et~al.}]{wei2022chain}
Jason Wei, Xuezhi Wang, Dale Schuurmans, Maarten Bosma, Fei Xia, Ed~Chi, Quoc~V Le, Denny Zhou, et~al. 2022.
\newblock Chain-of-thought prompting elicits reasoning in large language models.
\newblock \emph{Advances in neural information processing systems}, 35:24824--24837.

\bibitem[{Yao et~al.(2023{\natexlab{a}})Yao, Yu, Zhao, Shafran, Griffiths, Cao, and Narasimhan}]{yao2023tree}
Shunyu Yao, Dian Yu, Jeffrey Zhao, Izhak Shafran, Tom Griffiths, Yuan Cao, and Karthik Narasimhan. 2023{\natexlab{a}}.
\newblock Tree of thoughts: Deliberate problem solving with large language models.
\newblock \emph{Advances in neural information processing systems}, 36:11809--11822.

\bibitem[{Yao et~al.(2023{\natexlab{b}})Yao, Zhao, Yu, Du, Shafran, Narasimhan, and Cao}]{yao2023react}
Shunyu Yao, Jeffrey Zhao, Dian Yu, Nan Du, Izhak Shafran, Karthik Narasimhan, and Yuan Cao. 2023{\natexlab{b}}.
\newblock React: Synergizing reasoning and acting in language models.
\newblock In \emph{International Conference on Learning Representations (ICLR)}.

\bibitem[{Zhang et~al.(2019{\natexlab{a}})Zhang, Yu, Jiao, Xing, El~Ghaoui, and Jordan}]{zhang2019theoretically}
Hongyang Zhang, Yaodong Yu, Jiantao Jiao, Eric Xing, Laurent El~Ghaoui, and Michael Jordan. 2019{\natexlab{a}}.
\newblock Theoretically principled trade-off between robustness and accuracy.
\newblock In \emph{International conference on machine learning}, pages 7472--7482. PMLR.

\bibitem[{Zhang et~al.(2019{\natexlab{b}})Zhang, Kishore, Wu, Weinberger, and Artzi}]{zhang2019bertscore}
Tianyi Zhang, Varsha Kishore, Felix Wu, Kilian~Q Weinberger, and Yoav Artzi. 2019{\natexlab{b}}.
\newblock Bertscore: Evaluating text generation with bert.
\newblock \emph{arXiv preprint arXiv:1904.09675}.

\bibitem[{Zhao et~al.(2024)Zhao, Chen, Yang, Liu, Deng, Cai, Wang, Yin, and Du}]{zhao2024explainability}
Haiyan Zhao, Hanjie Chen, Fan Yang, Ninghao Liu, Huiqi Deng, Hengyi Cai, Shuaiqiang Wang, Dawei Yin, and Mengnan Du. 2024.
\newblock Explainability for large language models: A survey.
\newblock \emph{ACM Transactions on Intelligent Systems and Technology}, 15(2):1--38.

\end{thebibliography}

\end{document}